\documentclass[reqno]{amsart}
\numberwithin{equation}{section}
\usepackage{ewald}

\newcommand\pen[1]{(#1){}^{+}} 

\newcommand\Hess{D^{2}\ell\,\rvert_{f_{t}(x)}} 
\newcommand\Hessg{D^{2}\ell \,\rvert_{z}}
\newcommand\HessC{D^{2}\mathcal{C}\,\rvert_{\underline{x}}} 
\let\ran\relax
\DeclareMathOperator{\ran}{range}

\title[Gradient flow in parameter space is equivalent to
interpolation in output space]{Gradient flow in parameter space is equivalent to
linear interpolation in output space}

\author{Thomas Chen}
\address[T. Chen]{Department of Mathematics, University of Texas at Austin, Austin TX
78712, USA} \email{tc@math.utexas.edu}
\author{Patrícia Mu\~{n}oz Ewald}
\address[P. M. Ewald]{Department of Mathematics, University of Texas at Austin,
Austin TX 78712, USA} 
\email{ewald@utexas.edu}

\definecolor{burntorange}{HTML}{BF5700}
\definecolor{UTblue}{HTML}{00A9B7}
\definecolor{bluebonnet}{HTML}{005F86}
\hypersetup{
    pdfauthor={Patricia Munoz Ewald, Thomas Chen}, 
    pdftitle={Gradient flow in parameter space is equivalent to interpolation in output
    space},
    linktoc=all,     
    colorlinks=true, 
    linkcolor=bluebonnet, 
    citecolor=burntorange,
    urlcolor=bluebonnet, 
}

\begin{document}

\maketitle

\begin{abstract}
    We prove that the standard gradient flow in parameter space that underlies many training
    algorithms in deep learning can be continuously deformed into an
    adapted gradient flow which yields (constrained) Euclidean gradient flow in output
    space. 
    Moreover, for the $L^{2}$ loss,  if the Jacobian of the outputs with respect to the
    parameters is full rank (for fixed training data), then the time variable can be
    reparametrized so that the resulting flow is simply linear interpolation, and a global
    minimum can be achieved.  
    For the cross-entropy loss, under the same rank condition and assuming the labels have
    positive components, we derive an explicit formula for the unique global minimum.
\end{abstract}

\section{Introduction}

At the core of most algorithms currently used for training neural networks is gradient
descent, 
whose theoretical counterpart in the continuum limit is gradient flow. 
This flow is defined in a space of parameters $\mathbb{R}_{\underline{\theta}}^{K}$ with
respect to a cost function defined in $\mathbb{R}_{\underline{x}}^{QN}$.
The cost is non-convex as a function of the parameters, and
standard gradient flow in $\mathbb{R}_{\underline{\theta}}^{K}$ might not converge to a
global minimum. Even changing perspective to $\mathbb{R}_{\underline{x}}^{QN}$, where the
cost can be chosen to be convex, the resulting associated flow is complicated. 

In \cite{chen23}, it was discussed in detail how the choice of the Riemannian structure in
$\mathbb{R}_{\underline{\theta}}^{K}$
influences the training dynamics.  In this context, a
modified gradient flow that induces the Euclidean gradient flow in output space
$\mathbb{R}_{\underline{x}}^{QN}$ was
introduced and contrasted with the standard Euclidean gradient flow in parameter space;
it was shown that both flows exhibit the same critical sets. In the work at hand, we
extend these results by proving that those two flows are in fact homotopy equivalent to
one another (Theorem \ref{thm-homotopy}). 
Moreover, for the $L^{2}$ cost, we prove that if the Jacobian matrix exhibits no rank
loss, then the Euclidean flow in output space is reparametrization equivalent to linear
interpolation with respect
to a suitable time variable (Proposition \ref{prp-lindyn}). When there is rank loss, an
expression for the deviation from linear interpolation is given (Proposition
\ref{prp-rkloss}). For the cross-entropy loss, 
we prove that output space fibers into invariant manifolds over which Euclidean
gradient flow converges to a unique global minimum (Proposition \ref{prop:CE}).

These results underscore the central role of the rank of the Jacobian (or equivalently,
the neural tangent kernel) in shaping optimization dynamics. In particular,
if the gradient flow for a convex loss in output space reaches a fixed point, then at least one
of the following must be true: (i) the solution interpolates the data, or (ii) the
Jacobian is not injective.
This becomes evident when considering the constrained Euclidean gradient flow in output
space, as is done for the $L^{2}$ loss in section \ref{sec:repar} and for the
cross-entropy loss in section \ref{sec:CE}.

In section \ref{sec:applications}, we provide some applications: First, we note that 
instead of modifying the metric in $\mathbb{R}_{\underline{\theta}}^{K}$, one could
prescribe a convenient path in $\mathbb{R}_{\underline{x}}^{QN}$ and then find an
associated path in parameter space. Next, we show that neural collapse occurs in the
output layer for the trivialized dynamics studied in \ref{sec:repar} (Corollary
\ref{cor-neurcollapse}).
Finally, we rephrase Theorem \ref{thm-homotopy} in terms of the neural tangent kernel.

We briefly comment on related work in section \ref{sec:relatedwork}.  A short appendix on
pseudoinverses for rank-deficient matrices is provided as well.

\section{Main results}
\label{sec:main}

Consider a family of functions
\begin{align}
    f_{\underline{\theta}}: \mathbb{R}^{M} \to \mathbb{R}^{Q}
\end{align}
parametrized by a vector of parameters $\underline{\theta} \in \mathbb{R}^{K}$. Suppose we
wish to optimize $f_{\underline{\theta}}$ with respect to a given cost function
$\mathcal{C}:\mathbb{R}^{K} \to \mathbb{R}$. A natural approach is to pick a
starting point
$\underline{\theta}_{0} \in \mathbb{R}^{K}$ and construct a path
\begin{align}
    \underline{\theta}(s):[0, \infty) \to \mathbb{R}^{K}
\end{align}
via the gradient flow
\begin{align}
    \label{SGF}
    \partial_{s}\underline{\theta}(s) &= - \nabla_{ \underline{\theta }} \mathcal{C},
    \nonumber \\
    \underline{\theta}(0) &= \underline{\theta}_{0}.
\end{align}
We will refer to \eqref{SGF} as the \emph{standard gradient flow}.

When $f_{\underline{\theta}}$ is a neural network, the optimization problem has the
following structure: Consider a set $\mathcal{X}_{0} \subset \mathbb{R}^{M}$ (the training
data), such that $\abs{\mathcal{X}_{0}} = N$. We can form a data matrix $X_0 \in
\mathbb{R}^{M\times N}$ by making the vectors in $\mathcal{X}_{0}$ into the columns of
$X_0$.
Then we can consider a function
\begin{align}
    f( \underline{\theta}, X_0): \mathbb{R}^{K} \times \mathbb{R}^{M\times N} \to
    \mathbb{R}^{Q \times N},
\end{align}
defined by its action on each column of $X_0$: for $i=1, \cdots, N$, 
\begin{align}
    f(\underline{\theta}, X_0)_{i} = f_{\underline{\theta}}( (X_0)_{i}) \in
    \mathbb{R}^{Q}.
\end{align}
In the context of supervised learning, each data point $x \in \mathcal{X}_{0}$ is
associated to a desired output $y(x) \in \mathbb{R}^{Q}$, and so from a data matrix $X_0$
we can form  the corresponding matrix of labels, $Y \in \mathbb{R}^{Q\times N}$, 
\begin{align}
    Y_{i} = y( (X_0)_{i}).
\end{align}
The total cost $\mathcal{C}$ used for training is usually a sum of errors for each data
point,
\begin{align}
    \label{loss}
    \mathcal{C}(\underline{\theta}) = \sum_{x \in \mathcal{X}_{0}} \ell(f(
    \underline{\theta}, x), y(x)),
\end{align}
where the chosen loss function $\ell:\mathbb{R}^{Q}\times \mathbb{R}^{Q} \to \mathbb{R}$
depends on the parameters only through $f(\underline{\theta}, x).$

For fixed training data $X_0 = \left[ x_1 \cdots x_{j} \cdots x_N \right]$, we can also
consider the function
\begin{align}
    \underline{x}&: \mathbb{R}^{K} \to \mathbb{R}^{QN}
\end{align}
from parameter space to output space defined by
\begin{align}
    \label{xund}
    \underline{x}( \underline{\theta}) &:= ( f_{\underline{\theta}}(x_1) ^{T}, \cdots,
    f_{\underline{\theta}} (x_{N}) ^{T}) ^{T}.
\end{align}
We define the correspondent vector of labels $\underline{y}\in \mathbb{R}^{QN}$,
\begin{align}
    \underline{y} := (y(x_1)^{T}, \cdots, y(x_{N})^{T})^{T}.
\end{align}
In this case the cost can be redefined as
$\mathcal{C}(\underline{x}( \underline{\theta}))$, and only depends on the parameters
through its dependence on $\underline{x}( \underline{\theta})$.
The Jacobian $D[\underline{\theta}] \in \mathbb{R}^{QN \times K}$ is
\begin{align}
    \label{jacobian}
    \left(D[\underline{\theta}]\right)_{jk} &:= \frac{\partial
    \underline{x}_{j}[\underline{\theta}]}{\partial \theta_{k}},
\end{align}
and we assume from now on that the partial derivatives in \eqref{jacobian} exist and are
Lipschitz continuous for all $\underline{\theta}\in \mathbb{R}^{K}$.
We will often abbreviate $D = D[\underline{\theta}]$.

For  $\underline{\theta}(s) \in \mathbb{R}^{K}$, we can
consider the associated path
\begin{align}
    \underline{x}(s):[0,\infty ) \to \mathbb{R}^{QN}
\end{align}
given by
\begin{align}
    \underline{x}(s) &:= \underline{x}(\underline{\theta}(s))
\end{align}
from \eqref{xund}.
The associated Jacobian is now also time dependent, and we will write $D(s) :=
D[\underline{\theta}(s)]$, or simply $D$ for convenience. The chain rule gives
\begin{align}
    \label{chainrule}
    \partial_{s} \underline{x}(s) = D(s) \, \partial_{s}\underline{\theta}(s)
\end{align}
and
\begin{align}
    \label{chainrulecosts}
    \nabla_{\underline{\theta}} \mathcal{C} = D^{T} \nabla_{\underline{x}}\mathcal{C}.
\end{align}

\begin{remark} \label{theremark}
    It is clear that 
    \begin{align}
        \nabla_{\underline{x}( \underline{\theta})}
            \mathcal{C} = 0 \implies \nabla_{\underline{\theta}} \mathcal{C} = 0,
    \end{align}
    but the converse does not hold in general. 
    It is known that
    if $\rk D[\underline{\theta}] = QN = \min\{K,QN\}$, then
    $\nabla_{\underline{x}}\mathcal{C} = 0 \iff
    \nabla_{\underline{\theta}}\mathcal{C} = 0$.
    If $\mathcal{C}$ is the squared loss, then
    $\nabla_{\underline{x}}\mathcal{C}=0$ corresponds to a global minimum with zero loss.
    See e.g. \cite{tragerkohnbruna19,cooper21, chen23, karhadkaretal24, chenewald23gd}.

    More generally, consider a cost function for which the only minima in output space are
    global minima where the model has interpolated the data, $f(
    \underline{\theta}_{*}, x) = y(x)$ for all $x \in \mathcal{X}_{0}$; this is the case for
    the $L^{2}$ and cross-entropy losses. Then for an overparametrized model ($K\geq QN$),
    we can read the following from \eqref{chainrulecosts}: If the corresponding gradient
    flow reaches a fixed point, then at least one of the following holds,
    \begin{itemize}
        \item the model interpolates the data, or
        \item $D$ is not full-rank.
    \end{itemize}
    In other words, if the interpolating solution is not attainable, gradient flow will only
    converge when $D$ loses rank.
\end{remark}

\noindent\textbf{Notation}. Given a matrix $A \in
\mathbb{R}^{m \times n}$, we denote by $\pen{A}$ its pseudoinverse, or Moore--Penrose
inverse; see appendix
\ref{appendix} for a definition and useful properties. 
Moreover, given a subspace $V \subset \mathbb{R}^{d}$, we denote by $\mathcal{P}_{V}$ the
orthogonal projection onto $V$, such that $\mathcal{P}_{V} = \mathcal{P}_{V}^{2} =
\mathcal{P}_{V}^{T}$. 
In this work, $V$ will usually be the range (or image) of a linear
operator.

\subsection{Adapted gradient flow} 
By a simple computation using \eqref{chainrule} and \eqref{chainrulecosts}, if
$\underline{\theta}(s)$ satisfies the standard gradient flow \eqref{SGF}, then
\begin{align}
    \label{stdOutGF}
    \partial_{s}\underline{x} = -DD^{T} \nabla_{\underline{x}}\mathcal{C}.
\end{align}
We will refer to \eqref{stdOutGF} as \emph{standard gradient flow in output space}.
Suppose instead we wished to modify 
$\underline{\theta}(s)$ so that the associated path $\underline{x}(s)$ satisfies the
Euclidean gradient flow. We recall here results in previous work by one of the authors, in
the overparametrized case:
\begin{lemma}[\cite{chen23}]
    Let $K \geq QN$, and let $\underline{x}(s)=\underline{x}( \underline{\theta}(s))$
    be defined as in \eqref{xund}. 
    Assume $\underline{x}( \underline{\theta})$ has Lipschitz continuous
    derivatives.
    When $D$ is full rank, setting
    \begin{align}
        \partial_{s}\underline{\theta }(s) = - \pen{D^{T}D} \nabla_{\underline{\theta
        }}\mathcal{C}
    \end{align}
    yields
    \begin{align}
        \partial_{s}\underline{x}(s) = - \nabla_{\underline{x}}\mathcal{C}.
    \end{align}
    If we allow for $\rk (D) \leq QN$, then letting
    \begin{align}
        \partial_{s}\underline{\theta }(s) = -D^{T} \psi,
    \end{align}
    for $\psi$ satisfying
    \begin{align}
        \label{psieq}
        D D^{T}\psi = \mathcal{P}_{\ran(DD^{T})} \nabla_{\underline{x}} \mathcal{C},
    \end{align}
    results in
    \begin{align}
        \partial_{s}\underline{x}(s) = - \mathcal{P}_{\ran(DD^{T})}
        \nabla_{\underline{x}}\mathcal{C} .
    \end{align}
\end{lemma}

We will now condense these two separate cases into the same. First, note that 
\begin{align}
    \tilde{\psi} := D \pen{D} \pen{DD^{T}} \pen{D^{T}} \nabla_{\underline{\theta}}
    \mathcal{C} 
\end{align}
satisfies \eqref{psieq}, as the following computation shows:
\begin{align}
    D D^{T} \tilde{\psi} &=
    D \underbrace{D^{T} D \pen{D}}_{\text{$\stackrel{\eqref{identity1}}{=}  D^{T}$}}
    \pen{DD^{T}} \pen{D^{T}} \nabla_{\underline{\theta}} \mathcal{C} \nonumber \\
    &= D D^{T} \pen{DD^{T}} \pen{D^{T}} D^{T} \nabla_{\underline{x}} \mathcal{C} \nonumber
    \\ 
    &= D D^{T} \pen{D^{T}} \underbrace{\pen{D} \pen{D^{T}}
    D^{T}}_{\text{$\stackrel{\eqref{identity1}}{=}  \pen{D} $}}
    \nabla_{\underline{x}} \mathcal{C} \\
    &= D D^{T} \pen{D D^{T}} \nabla_{\underline{x}} \mathcal{C} \nonumber \\
    &= \mathcal{P}_{\ran(DD^{T})} \nabla_{\underline{x}} \mathcal{C}. \nonumber 
\end{align}
Moreover, we can rewrite
\begin{align}
    - \pen{D^{T}D} \nabla_{\underline{\theta
    }}\mathcal{C} &= -\pen{D^{T}D} (D^{T}D) \pen{D^{T} D} \nabla_{\underline{\theta}}
    \mathcal{C} \nonumber \\
                                         &= - (D^{T} D) \pen{D^{T}D} \pen{D^{T} D}
                                         \nabla_{\underline{\theta}} \mathcal{C} \nonumber
                                         \\ 
                                         &= - D^{T} \left[ D \pen{D} \pen{D^{T}} \pen{D}
                                         \pen{D^{T}} \right] \nabla_{\underline{\theta}}
                                         \mathcal{C} \\
                                         &= - D^{T} \left[ D \pen{D} \pen{DD^{T}} 
                                         \pen{D^{T}} \right] \nabla_{\underline{\theta}}
                                         \mathcal{C}, \nonumber \\
                                         &= -D^{T}\tilde{\psi} \nonumber.
\end{align}
Therefore in general, for $K \geq QN$ and $\rk(D) \leq QN$,
\begin{align}
    \label{adaptedGF}
    \partial_{s}\underline{\theta } = - \pen{D^{T}D} \nabla_{\underline{\theta
    }}\mathcal{C}
\end{align}
implies
\begin{align}
    \label{projectedEucGF}
    \partial_{s}\underline{x} = - \mathcal{P}_{\ran(DD^{T})}
    \nabla_{\underline{x}}\mathcal{C} = - \mathcal{P}_{\ran(D)}
    \nabla_{\underline{x}}\mathcal{C}.
\end{align}
We will refer to \eqref{adaptedGF} as \emph{adapted gradient flow} in
$\mathbb{R}_{\underline{\theta}}^{K}$ and \eqref{projectedEucGF} as
\emph{constrained Euclidean gradient flow} in
$\mathbb{R}_{\underline{x}}^{QN}$.\footnote{Note that this is not the same as the
    Projected Gradient Descent algorithm for optimization on a feasible set. In
\eqref{projectedEucGF} the \emph{direction} of the flow is constrained, not the value of
$\underline{x}$.}

It was further shown in \cite[Theorem 4.2]{chen23} that the standard gradient flow
\eqref{SGF} and the adapted gradient flow \eqref{adaptedGF} have the same equilibrium
points. 
We extend this to a family of interpolating gradient fields.

%

\begin{thm}
    \label{thm-homotopy}
    Let $\mathcal{C}:\mathbb{R}^{K}\to \mathbb{R}$ depend on $\underline{\theta}\in
    \mathbb{R}^{K}$ only through
    $\underline{x}(
    \underline{\theta})$, for $\underline{x}:\mathbb{R}^{K} \to
    \mathbb{R}^{QN}$ 
        a function with Lipschitz continuous derivatives
    defined as in \eqref{xund}, and let $D[\underline{\theta}] \in
    \mathbb{R}^{QN \times K}$ be the
    Jacobian defined in \eqref{jacobian}.  There is a one-parameter family of vector
    fields in $\mathbb{R}^{K}$ interpolating between
    \begin{align}
        - \nabla_{\underline{\theta}} \mathcal{C}
    \end{align}
    and
    \begin{align}
        -\pen{D^{T}D} \nabla_{\underline{\theta}} \mathcal{C};
    \end{align}
    this interpolation preserves singularities.

    Moreover, this induces an interpolation
    between the vector fields for the standard gradient flow in output space
    \eqref{stdOutGF} and the constrained Euclidean gradient flow \eqref{projectedEucGF}
    which preserves the equilibrium points of these flows.
\end{thm}

This holds for any loss function $\ell$ in \eqref{loss}, and to any neural
network whose activation function has Lipschitz continuous derivatives. 
Commonly used non-smooth activation functions, such as ReLU, can be smoothed to satisfy
this condition, although the resulting gradient flow might depend on the smoothing.

\begin{proof}
    Define the family of vector fields 
    \begin{align}
        V_{\underline{\theta}, \alpha} := -  A_{\underline{\theta},\alpha}
        \nabla_{\underline{\theta}}\mathcal{C}, \quad \alpha \in [0,1],
    \end{align} 
    for
    \begin{align} 
        A_{\underline{\theta}, \alpha} := \left( \alpha \pen{D^{T}D} + (1-\alpha )
        \Id_{K\times K} \right).
    \end{align}
    Then
    \begin{align}
        V_{\underline{\theta}, 0} = - \nabla_{\underline{\theta}} \mathcal{C} 
    \end{align}
    and
    \begin{align}
        V_{\underline{\theta}, 1} = - \pen{D^{T}D} \nabla_{\underline{\theta}}
        \mathcal{C}.
    \end{align}
    
    For $\alpha \in [0,1)$, note that $A_{\underline{\theta}, \alpha} > 0$ for all
    $\underline{\theta}$. For $\alpha =1$, writing $\mathcal{P}_{\ran(D^{T})} = \pen{D}D$, 
    we have 
    \begin{align}
        \nabla_{\underline{\theta}} \mathcal{C} =
        \pen{D}D \nabla_{\underline{\theta}} \mathcal{C},
    \end{align}
    since \eqref{chainrulecosts} shows $\nabla_{\underline{\theta}}\mathcal{C} \in
    \ran(D^{T})$, and also
    \begin{align}
        \ker\pen{D^{T}D} = \ker (D^{T}D)^{T} = \ker(D^{T}D) = \ker(\pen{D}D),
    \end{align}
    where we used \eqref{kerproperty} and Lemma
    \ref{appendix:lemma}.
    Thus, 
    $V_{\underline{\theta},\alpha}=0$ if, and only if, $ \nabla_{\underline{\theta}}
    \mathcal{C} = 0$, and 
    the vector fields $V_{\underline{\theta},\alpha}$ are singular at the same
    points $\underline{\theta}\in \mathbb{R}^{K}$, for all $\alpha \in [0,1]$.

    Suppose now that
    \begin{align}
        \partial_{s}\underline{\theta}(s) = V_{\underline{\theta}(s), \alpha},
    \end{align}
    so that for $\underline{x}(s) = \underline{x}(\underline{\theta}(s))$,
    \begin{align} 
        \partial_{s}\underline{x}(s) = D[\underline{\theta}(s)]
        V_{\underline{\theta}(s),\alpha}.
    \end{align}
    We wish to write $\partial_{s}\underline{x}$ in terms of
    $\nabla_{\underline{x}}\mathcal{C}$, so we compute
    \begin{align}
        D \, V_{\underline{\theta}, \alpha } 
            &= - D \left( \alpha \pen{D^{T}D} + (1-\alpha ) \Id_{K\times K} \right)
            \nabla_{\underline{\theta}}\mathcal{C} \nonumber \\
            &= - \left(\alpha D \pen{D^{T}D}  + (1-\alpha) D  \right)
            D^{T} \nabla_{\underline{x}}\mathcal{C} \nonumber \\
            &= - \left(\alpha D \pen{D^{T}D} D^{T}  + (1-\alpha) D D^{T}
            \right)  \nabla_{\underline{x}}\mathcal{C} \\
            &\stackrel{(*)}{=}  - \left(\alpha \mathcal{P}_{\ran(DD^{T})} + (1-\alpha )
            DD^{T}\right) \nabla_{\underline{x}}\mathcal{C} \nonumber \\
            &= - \left(\alpha \Id_{QN\times QN} + (1-\alpha ) DD^{T}\right)
            \mathcal{P}_{\ran(DD^{T})} \nabla_{\underline{x}}\mathcal{C},  \nonumber
    \end{align}
    where $(*)$ follows from \eqref{basicproperty3}, \eqref{projector1} and
    \eqref{identity1}.
    Therefore,
    \begin{align}
        \label{x-dynamics}
        \partial_{s}\underline{x}(s) = V_{\underline{x}(s),\alpha}
    \end{align}
    for vector fields in $\mathbb{R}^{QN}$ defined along the path $\underline{x}(s)$,
    \begin{align}
        V_{\underline{x}(s), \alpha} := - \left( \alpha \Id_{QN \times QN} + (1-\alpha)
        D(s) D(s)^{T}\right) \mathcal{P}_{\ran(D(s)D(s)^{T})} \nabla_{\underline{x}}
        \mathcal{C}.  
    \end{align}
    
    Observe that 
    \begin{align}
        V_{\underline{\theta}(s),0} = - \nabla_{\underline{\theta}} \mathcal{C}
    \end{align}
    and
    \begin{align} 
        V_{\underline{x}(s),1} = - \mathcal{P}_{\ran(DD^{T})}
        \nabla_{\underline{x}} \mathcal{C},
    \end{align}
    so that in fact $V_{\underline{x}(s),\alpha}, \,\alpha\in [0,1],$ interpolates between
    the standard dynamics given by \eqref{SGF} at $\alpha =0$, and constrained Euclidean
    gradient flow \eqref{projectedEucGF} for $\underline{x}(s)$ at $\alpha =1$.
    Moreover, for $\alpha \in (0,1]$,
    $(\alpha \Id_{QN\times QN} + (1-\alpha )DD^{T})$ is positive-definite, 
    and for $\alpha =0$, $DD^{T}$ is injective in the range of
    $\mathcal{P}_{\ran(DD^{T})}$, so
    \begin{align}
         ( \alpha \Id + (1-\alpha )DD^{T}) \mathcal{P}_{\ran(DD^{T})}
        \nabla_{\underline{x}} \mathcal{C} = 0 \text{ if, and only if, }
        \mathcal{P}_{\ran(DD^{T})}\nabla_{\underline{x}}\mathcal{C}=0.
    \end{align}
    Thus, the equilibrium points of \eqref{x-dynamics} are the same for all
    $\alpha\in[0,1]$.
\end{proof}

\vspace{1ex}
We add the following remarks highlighting the geometric interpretation of the above
results.  If $DD^T$ has full rank $QN$, then $V_{\underline{x}(s), \alpha}$ is the
gradient vector field with respect to the Riemannian metric on $T\mathbb{R}^{QN}$ with
tensor 
\begin{align}
	\left( \alpha \Id_{QN \times QN} + (1-\alpha) D
        D^{T}\right)^{-1} \,.
\end{align}
This means that $V_{\underline{x}(s), \alpha}$ can equivalently be considered as the
gradient fields obtained from the family of Riemannian structures
interpolating between the Euclidean structure on $T\mathbb{R}^{QN}$, and the metric
structure induced by the Euclidean structure on parameter space.

If $DD^T$ is rank-deficient, $\rk(DD^T)<QN$, we let $\mathcal{V}\subset T\mathbb{R}^{QN}$
denote the vector subbundle whose fibers are given by the range of $DD^{T}$. In this
situation,  $V_{\underline{x}(s), \alpha}$ is the gradient with respect to the bundle
metric $h_\alpha$ on $\mathcal{V}$ with tensor 
\begin{align}
	[h_\alpha]=\left( \alpha \Id_{QN \times QN} + (1-\alpha) D
        D^{T}\right)^{-1}\Big|_{\ran(DD^T[\underline{\theta}])} \,.
\end{align}
The triple $(\mathbb{R}^{QN},\mathcal{V},h_\alpha)$ defines a sub-Riemannian manifold with
a family of bundle metrics for $\alpha\in[0,1]$; see \cite{chen23} for a related
discussion and further details.

\subsection{Reparametrization for the \texorpdfstring{$L^{2}$}{L^2} loss}  
\label{sec:repar}
Assuming $\rk{D}=QN \leq K$, the adapted
gradient flow in output space resulting from \eqref{adaptedGF} is simply Euclidean
gradient flow, 
\begin{align}
    \label{eucGF}
    \partial_{s}\underline{x}(s) &= - \nabla_{\underline{x}} \mathcal{C}(\underline{x}(s)).
\end{align}
For 
\begin{align}
    \label{squaredloss}
    \mathcal{C}(\underline{x}(\underline{\theta})) = \frac{1}{2N} \abs{\underline{x}(
        \underline{\theta}) - \underline{y}}_{\mathbb{R}^{QN}}^{2}
\end{align}
we have
\begin{align}
    \partial_{s}\underline{x} = \frac{1}{N} (\underline{x}(s) - \underline{y}),
\end{align}
which is solved by
\begin{align}
    \label{eucGFsol}
    \underline{x}(s) = \underline{y} + e^{-\frac{s}{N}}(\underline{x}(0) - \underline{y}),
\end{align}
and we see that $\underline{x}(s) \to \underline{y}$ as $s\to \infty$.

\begin{remark}
    \label{remark:interpolation}
    This implies that $\underline{y}$ is in the image of $\underline{x}$, which
    is not necessarily the case in general. Recall from Remark \ref{theremark} that if the
    interpolating solution is not attainable, then $D$ must lose rank to reach a fixed
    point. Equation \eqref{eucGFsol} only holds
    if $D(t)$ is full rank for all $t\in [0,s]$ and the initialization $\underline{\theta}(0)$.
\end{remark}

Suppose instead that we wanted to impose a linear interpolation
\begin{align}
    \label{linearinterpol}
    \underline{\hat{x}}(t) &= \underline{y} + (1-t)(\underline{x}_{0} - \underline{y}).
\end{align}
Then letting
\begin{align}
    t &:= 1 - e^{-\frac{s}{N}}, \\
    s &= -N \ln (1-t),
\end{align}
and 
\begin{align}
    \label{timerepar}
    \underline{\tilde{x}}(t) := \underline{x}(-N \ln(1-t)),
\end{align}
for $\underline{x}(s)$ a solution to \eqref{eucGF},
we have
\begin{align}
    \label{xtilODE}
    \partial_{t} \underline{\tilde{x}}(t) &= \partial_{t}\underline{x}(-N \ln(1-t))
    \nonumber \\
                                          &= -\nabla_{\underline{x}}
                                          \mathcal{C}[\underline{\tilde{x}}(t)]
                                          \frac{N}{1-t} \nonumber \\
                                          &= - \frac{1}{1-t}( \underline{\tilde{x}}(t) -
                                          \underline{y}). 
\end{align}
Note that $\underline{\hat{x}}(t)$
solves \eqref{xtilODE}, as
\begin{align}
    \partial_{t} \underline{\hat{x}}(t) &= -( \underline{\hat{x}}_{0} - \underline{y})
    \nonumber \\
                                        &= -\frac{1}{1-t}(\underline{\hat{x}}(t) -
                                    \underline{y}).
\end{align}
Thus, we have proved the following:
\begin{prop}
    \label{prp-lindyn}
    If $\underline{x}(s)$ is a solution to Euclidean gradient flow \eqref{eucGF} with
    respect to the cost \eqref{squaredloss}, then 
    \begin{align}
        \underline{\tilde{x}}(t) := \underline{x}(-N \ln(1-t)) = \underline{y} +
        (1-t)(\underline{x}_{0} - \underline{y}),
    \end{align}
    and $\underline{\tilde{x}}(t) \to \underline{y}$ as $t \to 1$. In particular, this
    holds for  
    $\underline{x}(s) = \underline{x}( \underline{\theta}(s))$ defined as in \eqref{xund}
    when $\underline{\theta}(s)$ is a solution to the adapted gradient flow
    \eqref{adaptedGF} and  $\rk D[\underline{\theta}(s)] = QN \leq K$.
\end{prop}

On the other hand, if $\rk(D)<QN$, the reparametrized Euclidean gradient flow in output
space provides a concrete criterion whereby the effect of rank loss can be measured.

\begin{prop}\label{prp-rkloss}
    In general, the constrained Euclidean gradient flow in output space with time
    reparametrization as in \eqref{timerepar} satisfies
    \begin{align}\label{eq-xtilODE-rankloss}
		\partial_t \underline{\tilde{x}}(t) = -\frac{1}{1-t}
        \mathcal{P}_t(\underline{\tilde{x}}(t)-\underline{y}) \, ,
        \;\;\;\underline{\tilde{x}}(0)=\underline{x}_0 \, , \;\;\;
		t\in[0,1)\, ,
	\end{align}
    where $\underline{\tilde{x}}(t)=\underline{x}[\underline{\theta}(s(t))]$ and
    $\mathcal{P}_t:=\mathcal{P}_{\ran(DD^T)}[\underline{\theta}(s(t))]$.  When $\rk(D) <
    QN$, the deviation from linear interpolation is given by
    \begin{align}\label{eq-xtil-rankloss-dev}
        \underline{\tilde{x}}(t)- \big((1-t)\underline{x}_0 + t\underline{y}\big) =\int_0^t
        dt' \, \mathcal{U}_{t,t'} \, \frac{1-t}{1-t'} \, \mathcal{P}^\perp_{t'} \,
        (\underline{x}_0-\underline{y}),
    \end{align}
    where the linear propagator $\mathcal{U}_{t,t'}$ is determined by
    \begin{align}
    	\partial_t \mathcal{U}_{t,t'} =
    	\frac{1}{1-t} \, \mathcal{P}_t \, \mathcal{U}_{t,t'} \, ,
    	\;\;\;
    	\mathcal{U}_{t',t'} = \Id_{QN\times QN} \, ,
    	\;\;\;t,t'\in[0,1)\,.
    \end{align}
\end{prop}

\begin{proof}
The equation \eqref{eq-xtilODE-rankloss} is straightforwardly obtained in a similar way as \eqref{xtilODE}.

To prove \eqref{eq-xtil-rankloss-dev}, we write
\begin{align}\label{eq-xtil-rankloss-dev-1}
	\underline{\tilde{x}}(t) = (1-t)\underline{x}_0+t\underline{y}
	+(1-t)R(t)
\end{align}
so that
\begin{align}\label{eq-xtil-rankloss-dev-2}
	\partial_t\underline{\tilde{x}}(t) = -(\underline{x}_0-\underline{y})
	-R(t)+(1-t)\partial_t R(t) \,.
\end{align}
On the other hand,
\begin{align}
		\partial_t \underline{\tilde{x}}(t) &= -\frac{1}{1-t}
		\mathcal{P}_t(\underline{\tilde{x}}(t)-\underline{y})
		\nonumber\\
		&= -\frac{1}{1-t}(\underline{\tilde{x}}(t)-\underline{y})
		+\frac{1}{1-t}\mathcal{P}_t^\perp (\underline{\tilde{x}}(t)-\underline{y})
		\\
		&= -\frac{1}{1-t}\big( (1-t)\underline{x}_0+t\underline{y}-\underline{y}
		+(1-t)R(t)\big)
        +\frac{1}{1-t}\mathcal{P}_t^\perp
        \big((1-t)\underline{x}_0+t\underline{y}-\underline{y} +(1-t)R(t))\big)
		\nonumber\\
		&= -(\underline{x}_0-\underline{y})
		- R(t)
		+\mathcal{P}_t^\perp \big(\underline{x}_0-\underline{y}
		+ R(t)\big), \nonumber
\end{align}
where we used \eqref{eq-xtil-rankloss-dev-1} to pass to the third line. Comparing the last
line with \eqref{eq-xtil-rankloss-dev-2}, we find \begin{align}
	\partial_t R(t) = \frac1{1-t}\mathcal{P}_t^\perp R(t)
	+ \frac1{1-t}\mathcal{P}_t^\perp (\underline{x}_0-\underline{y}) \,.
\end{align}
Solving this matrix valued linear ODE for $R(t)$ 
using the Duhamel (or variation of
constants) formula (see e.g., \cite{teschlODE}), 
and substituting the resulting expression in
\eqref{eq-xtil-rankloss-dev-1} yields \eqref{eq-xtil-rankloss-dev}, as claimed.
\end{proof}

\subsection{Cross-entropy loss} \label{sec:CE}

For classification of data into $Q$ classes, the output $f_{\underline{\theta}}(x)$ and
label $y(x)$ associated to an input $x \in \mathcal{X}_{0}$ are vectors in
$\mathbb{R}^{Q}$ 
representing the
probability that $x$ belongs to each class, 
so that $\ f_{\underline{\theta}}(x)_{j}, y_{j} \in [0,1]$, $j=1,
\cdots, Q$, and $\sum_{j=1}^{Q} f_{\underline{\theta}}(x)_{j} = \sum_{j=1}^{Q} y_{j} = 1.$
The cross-entropy loss is
\begin{align}
    \ell (f(x),y) = - y \cdot \log(f(x)) = - \sum_{j=1}^{Q} y_{j} \log(f(x)_{j}),
\end{align}
where the $\log$ is taken component-wise on vectors.
In this case, 
the task becomes one of comparing the probability
distribution of the true labels with that of the predicted labels.
Prior work has explored the geometric properties of gradient flows on such probability
manifolds within the framework of information geometry, see e.g.
\cite{amari-informationgeometry, ayjost-informationgeometry, limontufar18}.  
Whether the ideas presented here can be meaningfully extended to this setting remains an
open question, which we leave for future investigation.

\vspace{1ex}
Let us consider instead a model whose output can be any vector in
$\mathbb{R}^{Q}$.
Let $\sigma$ be the softmax function, so that if $z\in \mathbb{R}^{Q}$, $\sigma(z)\in
\mathbb{R}^{Q}$ and 
\begin{align}
    \sigma(z)_{j} = \frac{e^{z_{j}}}{\sum_{i} e^{z_{i}}}, \quad j=1, \cdots, Q,
\end{align}
and define
\begin{align}
    \label{CEloss}
    \ell (f(x),y) = - \sum_{j=1}^{Q} y_{j} \log \sigma(f(x))_{j} = - y \cdot \log
    \sigma(f(x))
\end{align}
with $y_{j} \in [0,1]$ for $j=1, \cdots, Q$ and $\sum_{i=1}^{Q} y_{i} = 1$.
Then Theorem \ref{thm-homotopy} applies to 
\begin{align}
    \label{CEcost}
    \mathcal{C}( \underline{\theta}) = \sum_{x \in \mathcal{X}_{0}}
    \ell(f_{\underline{\theta}}(x), y(x)).
\end{align}

Assuming $\rk{D}=QN \leq K$, 
the gradient flow in output space resulting from \eqref{adaptedGF} is simply
Euclidean gradient flow, 
\begin{align}
    \label{eucGFCE}
    \partial_{t}\underline{x}(t) &= - \nabla_{\underline{x}} \mathcal{C}(\underline{x}(t)).
\end{align}
From now until the end of this subsection, we 
hide the dependence on the parameters $\underline{\theta} \in \mathbb{R}^{K}$ and
let $f_{t}(x) := f( \underline{\theta}(t), x)$.  We can compute, for some input $x \in
\mathcal{X}_{0}$ and associated label $y = y(x) \in \mathbb{R}^{Q}$, 
\begin{align}
    \label{gradCEcost}
    \partial_{f(x)_{k}} \mathcal{C} &=  \sum_{i \neq k} \left( y_{i}  \frac{\sum_{j}
    e^{f(x)_{j}}}{ e^{f(x)_{i}}} \frac{e^{f(x)_{i}} e^{f(x)_{k}}}{\left(\sum_{j}
    e^{f(x)_{j}}\right)^{2}} \right)
    - y_{k} \frac{\sum_{j} e^{f(x)_{j}}}{e^{f(x)_{k}}} \frac{\left(\sum_{j}
    e^{f(x)_{j}}\right) e^{f(x)_{k}} - e^{2 f(x)_{k}}}{\left(\sum_{j}
    e^{f(x)_{j}}\right)^{2}} \nonumber \\
                            &= \sum_{i \neq k} \left( y_{i} \frac{e^{f(x)_{k}}}{\sum_{j}
                            e^{f(x)_{j}}}\right) -
                            y_{k} \frac{\sum_{j} e^{f(x)_{j}} - e^{f(x)_{k}}}{\sum_{j}
                            e^{f(x)_{j}}} \nonumber \\
                            &= -y_{k} + \frac{e^{f(x)_{k}}}{\sum_{j}
                            e^{f(x)_{j}}} \sum_{i=1}^{Q} y_{i} \\
                            &= -y_{k} + \frac{e^{f(x)_{k}}}{\sum_{j}
                            e^{f(x)_{j}}}, \nonumber
\end{align}
and end up with a system of ordinary differential equations for each input: 
\begin{align}
    (f_{t}(x)_{j})' = - \partial_{f_{t}(x)_{j}} \mathcal{C} = y_{j} - 
    \frac{e^{f_{t}(x)_{j}}}{\sum_{i} e^{f_{t}(x)_{i}}}, \quad j=1, \cdots, Q,
\end{align}
or
\begin{align}
    \label{CE-ODE}
    f_{t}(x)' = - \nabla_{f_{t}(x)} \mathcal{C} = y - \sigma(f_{t}(x)).
\end{align}

We make two key observations: First, note that
\begin{align}
    \sum_{j=1}^{Q} f_{t}(x)_{j}' = \underbrace{\sum_{j=1}^{Q} y_{j}}_{=1} -
    \underbrace{\frac{\sum_{j=1}^{Q} e^{f_{t}(x)_{j}}}{\sum_{i=1}^{Q} e^{f_{t}(x)_{i}}}}_{=1} 
    = 0,
\end{align}
so 
\begin{align}
    \sum_{j=1}^{Q} f_{t}(x)_{j}(s)  = \sum_{j=1}^{Q} f_{0}(x)_{j} =: c.
\end{align}
Thus, even though we assumed that $f_{t}(x) \in \mathbb{R}^{Q}$ is not necessarily on the
probability simplex, so that $c \in \mathbb{R}$ can have any value, 
a solution to \eqref{CE-ODE} is still constrained to an affine hyperplane
\begin{align}
    \label{hyperplane}
    H_{c} := \left\{ z \in \mathbb{R}^{Q}: z \cdot u_{Q} = c \right\}
\end{align}
where $u_{Q} := (1, \cdots, 1)^{T} \in \mathbb{R}^{Q}$. 
This is in line with the fact that
the softmax function is invariant under addition of a scalar multiple of $u_{Q}$.

Second, we compute the Hessian of the loss \eqref{CEloss} for a single data point, 
\begin{align}
    \left[ D^{2} \ell \right]_{jk} &= \partial_{f_{t}(x)_{k}} \left( -y_{j} +
    \frac{e^{f_{t}(x)_{j}}}{\sum_{i} e^{f_{t}(x)_{i}}}\right) \nonumber \\
                                         &= \delta_{jk}\frac{e^{f_{t}(x)_{j}}}{\sum_{i}
                                         e^{f_{t}(x)_{i}}} -
                                         \frac{e^{f_{t}(x)_{j}}}{\sum_{i}
                                             e^{f_{t}(x)_{i}}}
                                             \frac{e^{f_{t}(x)_{k}}}{\sum_{i}
                                         e^{f_{t}(x)_{i}}},
\end{align}
so
\begin{align}
    \Hess
    = \diag(\sigma(f_{t}(x))) - \sigma(f_{t}(x)) \sigma(f_{t}(x))^{T} .
\end{align}
Observe that
\begin{align}
    \Hess \, u_{Q} = \sigma(f_{t}(x)) - \sigma(f_{t}(x))
    \underbrace{(\sigma(f_{t}(x)) \cdot u_{Q})}_{=
    \sum_{j} \sigma(f_{t}(x))_{j} \,=\, 1} = 0.
\end{align}

\begin{lemma}
    \label{lemma:hessian}
    Let $\Hessg \in \mathbb{R}^{Q\times Q}$ be the Hessian matrix of the loss
    \eqref{CEloss} evaluated at an arbitrary point $z \in \mathbb{R}^{Q}$,
    \begin{align}
        \Hessg = \diag(\sigma(z)) - \sigma(z)\sigma(z)^{T}.
    \end{align}
    For all $z \in \mathbb{R}^{Q}$,
    $\rk \Hessg = Q-1$ and $\Hessg$ is
    positive-semidefinite. 
\end{lemma}

\begin{proof}
    Fix $z\in \mathbb{R}^{Q}$.
    First, since $\sigma(z)_{j} > 0$ for all $j=1, \cdots, Q$,  we have $\rk
    \diag(\sigma(z)) = Q$. Moreover, $\rk (\sigma(z)
    \sigma(z)^{T}) = 1$, so 
    \begin{align} 
        \rk \Hessg \geq \rk \diag(\sigma(z)) - \rk(\sigma(z)
        \sigma(z)^{T}) = Q - 1.
    \end{align}
    On the other hand, 
    \begin{align}
        \Hessg u_{Q} = 0,
    \end{align}
    so $\dim (\ker \Hessg) \geq 1$.
    Therefore,
    \begin{align}
        Q - 1 \leq \rk \Hessg = Q - \dim(\ker \Hessg) \leq Q-1.
    \end{align}
    
    
    \vspace{1ex}
    To show $\Hessg \geq 0$, let $S := \diag(\sigma(z))$.
    Then for any $v\in \mathbb{R}^{Q}$, 
    \begin{align}
        \langle v, \Hessg v\rangle &= \langle v, S v\rangle - \abs{\langle v,
        \sigma(z)\rangle}^{2}.
    \end{align}
    Using Cauchy--Schwarz,
    \begin{align}
        \abs{\langle v,\sigma(z)\rangle}^{2} &= \abs{\langle S^{\frac{1}{2}}v,
        S^{-\frac{1}{2}} \sigma(z)\rangle}^{2} \\
                                             &\leq \langle v, Sv\rangle \langle
                                             \sigma(z),
                                             S^{-1} \sigma(z) \rangle.
    \end{align}
    Therefore
    \begin{align}
        \langle v,\Hessg v\rangle \geq \langle v, Sv\rangle\left(1 - \langle
        \sigma(z),S^{-1}\sigma(z)\rangle \right),
    \end{align}
    but
    \begin{align}
        S^{-1} \sigma(z) = u_{Q}
    \end{align}
    and 
    \begin{align}
        \langle \sigma(z), S^{-1} \sigma(z)\rangle =
        \sigma(z) \cdot u_{Q} = \sum_{j=1}^{Q} \sigma(z)_{j} = 1,
    \end{align}
    thus
    \begin{align}
        \langle v,\Hessg v\rangle \geq 0.
    \end{align}
\end{proof}

From this we conclude that $\Hessg$ is positive-definite when restricted to the
submanifolds $H_{c}$, so that for each training data point the cross-entropy loss
\eqref{CEloss} appears convex to a solution of the Euclidean gradient flow
\eqref{CE-ODE}. If there exists a critical point, it should correspond to a global
minimum.
Such an equilibrium point must satisfy
\begin{align}
    \sigma(f_{*}(x)) = y,
\end{align}
and this can only be attained if $y_{j} > 0$, for all $j=1, \cdots, Q$.\footnote{
While it is common to train classifiers using one-hot encoded labels, where $y\in
\mathbb{R}^{Q}$ are canonical
basis vectors, training with labels that assign positive probabilities to all classes
leads to better convergence properties, as we show in the present work. This scenario can
occur when certain training inputs do not clearly belong to a single class, or can be
forced by subjecting the labels to small perturbations.}
We have thus proved most of the following:

\begin{prop}
    \label{prop:CE}
    If $f_{t}(x)$ is a solution to the gradient flow \eqref{CE-ODE} with respect to the
    cost \eqref{CEcost}, then its image is restricted to an affine hyperplane,
    \begin{align}
        \sum_{j=1}^{Q} f_{t}(x)_{j} = c,
    \end{align}
    where $c = \sum_{j=1}^{Q} f_0(x)_{j}$.

    Moreover, if the
    associated label $y \in \mathbb{R}^{Q}$ has positive components, then 
    $f_{t}(x) \to f_{*,c}(x)$  as $t \to \infty$, with
    \begin{align}
        \label{f_c}
        f_{*,c}(x) = \log y + \frac{1}{Q}\left(\sum_{j=1}^{Q} (f_{0}(x)_{j} -
        \log y_{j}) \right)u_{Q},
    \end{align}
    where $\log y = (\log y_1, \cdots, \log y_{Q})^{T}$.
\end{prop}

\begin{remark}
    As in the previous section, this corresponds
    to an interpolating solution.  See Remarks \ref{theremark} and \ref{remark:interpolation}.
\end{remark}

\vspace{1ex}
Note that
\begin{align}
    \nabla_{\underline{x}(t)}\mathcal{C} = (\nabla_{f_{t}(x_1)}\ell(f_{t}(x_1), y(x_1))^{T},
    \cdots, \nabla_{f_{t}(x_{N})}\ell(f_{t}(x_{N}), y(x_{N}))^{T}) ^{T},
\end{align}
and as with the $L^{2}$ loss, the dynamics of training for each data point in the set of
training data
$\mathcal{X}_{0}$ is independent. Further, the Hessian of the total cost \eqref{CEcost} is
block-diagonal,
\begin{align}
    \HessC 
    = \diag( D^{2} \ell \rvert_{f_{t}(x_1)},
    \cdots, D^{2} \ell \rvert_{f_{t}(x_{N})}) \in \mathbb{R}^{QN \times QN}.
\end{align}
It is straightforward to check that $\HessC$ is positive-definite when restricted to a
product of hyperplanes $H_{c_1} \times \cdots \times H_{c_{N}} \subset \mathbb{R}^{QN}$.
Therefore, we may extend Proposition \ref{prop:CE} accordingly.

\begin{coro}
    \label{coro:CE}
    If $\underline{x}(t) = (f_{t}(x_1)^{T}, \cdots, f_{t}(x_{N})^{T})^{T} \in
    \mathbb{R}^{QN}$ is a solution to Euclidean gradient flow \eqref{eucGFCE} with respect
    to the cost \eqref{CEcost},
    then $\underline{x}(t)$ is constrained to a product of $N$ affine hyperplanes in
    $\mathbb{R}^{Q}$, defined by
    \begin{align}
        \sum_{j=1}^{Q} f_{t}(x_{n})_{j} = c_{n}, 
    \end{align}
    where 
    $c_{n} = \sum_{j=1}^{Q} f_{0}(x_{n})_{j}$, for
    $n=1, \cdots, N$.

    Moreover, if the
    associated label vector $\underline{y} \in \mathbb{R}^{QN}$ has positive components,
    then there exists a family of
    global minima $\underline{x}^{*}(C)$ parametrized by $C = (c_1, \cdots, c_{N}) \in
    \mathbb{R}^{N}$, and
    $\underline{x}(t)$ converges
    to a unique equilibrium 
    \begin{align}
        \underline{x}^{*}(C) &= (f_{*,c_1}(x_1)^{T}, \cdots, f_{*,c_{N}}(x_{N})^{T})^{T}
    \end{align}
    as $t \to \infty$, for $f_{*,c}$ defined in \eqref{f_c}. 
\end{coro}

\begin{proof}
    All that is left to prove Proposition \ref{prop:CE} and Corollary \ref{coro:CE} is to
    find a formula for $f_{*,c}(x)$.  Suppose that $f_{t}(x)$ is such that
    \begin{align}
        e^{f_{t}(x)_{j}} = \eta y_{j}, \quad j=1, \cdots, Q,
    \end{align}
    so that
    \begin{align}
        \sigma(f_{t}(x))_{j} = \frac{e^{f_{t}(x)_{j}}}{\sum_{i} e^{f_{t}(x)_{i}}} = \frac{\eta
        y_{j}}{\sum_{i} \eta y_{i}} = y_{j}.
    \end{align}
    This satisfies $f_{t}(x)' = 0$ for \eqref{CE-ODE}, and is thus an equilibrium
    point for the gradient flow. 
    Then
    \begin{align}
        f_{*,c}(x)_{j} = \log(\eta y_{j}) = \log(y_{j}) + \log(\eta), \quad j=1, \cdots, Q.
    \end{align}

    If
    \begin{align}
        \sum_{j=1}^{Q} f_{0}(x)_{j} = c,
    \end{align}
    then 
    \begin{align}
        c = \sum_{j=1}^{Q} f_{*,c}(x)_{j} =  \sum_{j=1}^{Q} \log y_{j} + Q \log(\eta)
    \end{align}
    and
    \begin{align}
       \log(\eta) = \frac{1}{Q} \left(c - \sum_{j} \log y_{j} \right),
    \end{align}
    from which we conclude
    \begin{align}
        f_{*,c}(x)_{j} &= \log y_{j} + \frac{1}{Q} \left(c - \sum_{i=1}^{Q} \log y_{i} \right)
        \nonumber \\
             &= \log y_{j} + \frac{1}{Q} \sum_{i=1}^{Q} (f_{0}(x)_{i} - \log y_{i})
    \end{align}
    for each $j=1, \cdots, Q.$
    Since $\Hess$ is positive-definite on the hyperplane $H_{c}$, this minimum is unique
    on $H_{c}$.

    Finally, as 
    \begin{align}
        \ell (f_{t}(x), y) = - y \cdot \log \sigma(f_{t}(x)),
    \end{align}
    we have
    \begin{align}
        \ell(f_{*,c}(x), y) = - y \cdot \log y > 0
    \end{align}
    which does not depend on the initial value $f_{0}(x)$. Therefore,
    $f_{*,c}(x)$ is a degenerate global minimum. 
\end{proof}
   
\begin{remark}
    For a fixed $x\in \mathcal{X}_{0}$ and two different sets of initial conditions,
    \begin{align}
        f_{*,c_1}(x) - f_{*,c_2}(x) = \frac{1}{Q}(c_2 - c_1) u_{Q}
    \end{align}
    or
    \begin{align}
        f_{*,c_1}(x) = f_{*,c_2}(x) + \frac{c_2 - c_1}{Q} u_{Q}.
    \end{align}
    This implies that the equilibrium manifold is a straight line passing through
    $f_{*,c}(x)$, for any fixed $c\in \mathbb{R}$. We conclude that $\mathbb{R}^{Q}$
    fibers into invariant manifolds $H_{c}$ for the adapted Euclidean gradient flow, and
    that the critical manifold $Crit(x) = \left\{z : \nabla_{z} \ell(z,y(x)) = 0
    \right\}$ is
    given by $\left\{f_{*,c}(x) + \lambda u_{Q}: c\in \mathbb{R}\text{ fixed}, \lambda\in
    \mathbb{R}\right\}$. Moreover, $Crit(x)$ and $H_{c}$ intersect transversally, for all
    $c\in \mathbb{R}.$
\end{remark}

\section{Applications} \label{sec:applications}

\subsection{Prescribed paths in output space}
\label{sec:newalgo}

In the previous section, we discussed how changes in $\partial_{s}\underline{\theta}(s)$
influence $\partial_{s}\underline{x}(s)$. It could be interesting to take a different point
of view and find $\underline{\theta}(s)$ from a given path $\underline{x}(s)$.
In general, from \eqref{chainrule},
\begin{align}
    \mathcal{P}_{\ran(D^{T})} \partial_{s}\underline{\theta} = \pen{D}
    \partial_{s}\underline{x},
\end{align}
and so assuming $\partial_{s}\underline{\theta} = \mathcal{P}_{\ran(D^{T})}
\partial_{s}\underline{\theta}$,\footnote{Note that this is satisfied  by the standard
    gradient flow $\partial_{s}\underline{\theta} = -\nabla_{\underline{\theta}}
\mathcal{C} = - D^{T} \nabla_{\underline{x}} \mathcal{C}.$ }
we can write
\begin{align}
    \label{prescribedDtheta}
    \partial_{s}\underline{\theta}(s) = \pen{D} \partial_{s}\underline{x}(s) .
\end{align}
Thus, we could prescribe a path 
\begin{align}
    \label{xpath}
    \underline{x}(s), &\quad s\in [0, T], \nonumber \\
    \partial_{s}\underline{x}(s) &\in \ran(D(s)), \\
    \underline{x}(T) &= \underline{y}, \nonumber
\end{align}
and then 
search for a corresponding
$\underline{\theta}_{*}$ which realizes
\begin{align}
    \underline{x}( \underline{\theta}_{*}) = \underline{x}(T)
\end{align}
by solving
\begin{align}
    \label{newalgo}
    \partial_{s}\underline{\theta}(s) &= \pen{D(s)} \partial_{s}\underline{x}(s),
    \nonumber \\ 
    \underline{\theta}(0) &= \underline{\theta}_{0},\\
    \underline{x}(s) &\text{ satisfies \eqref{xpath} with } \underline{x}(0) =
    \underline{x}( \underline{\theta}_{0}), \nonumber
\end{align}
for some initialization $\underline{\theta}_{0} \in \mathbb{R}^{K}$.
Observe that
\begin{align}
    \ran(D) = (\ker(D^{T}))^{\perp} = (\ker\pen{D})^{\perp},
\end{align}
and so all equilibrium points of \eqref{newalgo} satisfy $\partial_{s}\underline{x} = 0$.

As an example, 
motivated by the results in section \ref{sec:repar}, 
one could take $\underline{x}(s)$ to be the linear interpolation
\eqref{linearinterpol} and check if,
for a given
$\underline{\theta}_{0}$,
\begin{align}
    \label{xpath_interpol}
    (\underline{x}_{0} - \underline{y}) \in \ran{D(s)},
\end{align}
for all $s\in [0,T]$. 
\begin{remark}
    Whether or not a path $\underline{x}(s)$ satisfies the constraint in \eqref{xpath} (or
    \eqref{xpath_interpol} in the example above) could depend on the choice of
    $\underline{\theta}_{0}$. In general, properties of the Jacobian matrix
    $D[\underline{\theta}(s)]$ that determine the dynamics of $\underline{x}(s)$, such as
    rank and the subspace $\ran(DD^{T})$, might depend on the initialization even
    within a class $\left\{ \underline{\theta}_{0} : \underline{x}(
        \underline{\theta}_{0}) = \underline{x}_{0}\right\}$.
\end{remark}

\subsection{Final layer collapse} \label{sec:collapse}
We turn to the case of a classification task. 
Assume without any loss of generality that the components of
$\underline{\tilde{x}}$ are ordered by way of 
\begin{align}
	\underline{\tilde{x}}=(\dots\dots,\tilde{x}_{j,1}^T,\dots,
	\tilde{x}_{j,i_j}^T,\dots,\tilde{x}_{j,N_j}^T,\dots\dots)^T\in\mathbb{R}^{QN}
\end{align}
where each $\tilde{x}_{j,i_j}\in\mathbb{R}^Q$ belongs to the class labeled by $y_j \in
\mathbb{R}^{Q}$, for
$j=1,\dots,Q$, $i_j=1,\dots,N_j$, and $\sum_{j=1}^Q N_j=N$. Then, defining the class
averages 
\begin{align}
	\overline{\tilde{x}_j}(t):=\frac{1}{N_j}\sum_{i_j=1}^{N_j}\tilde{x}_{j,i_j}(t)
\end{align}
and the deviations
\begin{align}
	\Delta\tilde{x}_{j,i_j}(t):=\tilde{x}_{j,i_j}(t)-\overline{\tilde{x}_j}(t)\,,
\end{align}
the $L^2$ cost \eqref{squaredloss} decomposes into
\begin{align}
    \label{cost-decomp}
	\mathcal{C}[\underline{\tilde{x}}(t)]
	&=\frac1N\sum_{j=1}^Q\sum_{i_j}^{N_j}|\tilde{x}_{j,i_j}(t)-y_j|^2
	\nonumber\\
	&= \frac1N\sum_{j=1}^Q\sum_{i_j}^{N_j}
	\big(|\tilde{x}_{j,i_j}(t)
	-\overline{\tilde{x}_j}(t)|^2+|\overline{\tilde{x}_j}(t)-y_j|^2\big)
	\nonumber\\
	&= \frac1N\sum_{j=1}^Q\sum_{i_j}^{N_j}|\Delta\tilde{x}_{j,i_j}(t)|^2
	+\sum_{j=1}^Q|\overline{\tilde{x}_j}(t)-y_j|^2 \,.
\end{align}
This implies that zero loss training is achievable if and only if the class averages in
the output layer are matched to the reference outputs,
$\overline{\tilde{x}_j}(t)\rightarrow y_j$ for all $j=1,\dots,Q$, and all deviations
vanish, $\Delta\tilde{x}_{j,i_j}(t)\rightarrow 0$ for all $j=1,\dots,Q$,
$i_j=1,\dots,N_j$. The latter implies that the image of training data in
the output layer contracts to one point per class.

Thus, Proposition \ref{prp-lindyn} allows us to arrive at the following conclusion.

\begin{coro}\label{cor-neurcollapse}
    If $\rk(D) = QN$, then zero loss minimization is achieved, whereby the cluster
    averages converge to the reference outputs, 
    \begin{align}
	\lim_{t\rightarrow1^-}\overline{\tilde{x}_j}(t)=y_j,
    \end{align}
    and the deviations converge to zero,
    \begin{align}
    	\lim_{t\rightarrow1^-}\Delta{\tilde{x}}_{j,i_j}(t)=0 \,,
    \end{align}
    for all $j=1,\dots,Q$ and $i_j=1,\dots,N_j$.
\end{coro}

The decomposition of the cost function \eqref{cost-decomp} is related to the phenomenon
known as neural collapse (\cite{papyanhandonoho20, hanpapyandonoho22}), which takes
place on the penultimate layer of a neural network.
For work on the relationship between neural collapse and final layer collapse, see e.g.
\cite{ewojt22}.

\subsection{Tangent kernel} \label{sec:ntk}
A related point of view is to study the neural tangent kernel (NTK) introduced by
\cite{jacotetal18}. 
It can be written as
\begin{align}
    \Theta(x, y, \underline{\theta}) &= Df_{\underline{\theta}}(x)
    (Df_{\underline{\theta}}(y))^{T} \in \mathbb{R}^{Q\times Q},
\end{align}
for a function $f_{\underline{\theta}}:\mathbb{R}^{M}\to \mathbb{R}^{Q}$ and
$Df_{\underline{\theta}}(x)$ the Jacobian of $f_{\underline{\theta}}$ at $x\in
\mathbb{R}^{M}$.
Then for a set of training data $\mathcal{X}_0 = \{ x_1, \cdots, x_N\}$ and
$D[\underline{\theta}]$ defined as in section \ref{sec:main} we can write $DD^{T} \in
\mathbb{R}^{QN\times QN}$ as a
block matrix with blocks of size $Q\times Q$,
\begin{align}
    \label{DDT-ntk}
    (D[\underline{\theta}] D[\underline{\theta}]^{T})i,j &= 
        \Theta(x_i, x_j,\underline{\theta}), \quad i,j=1, \cdots, N.
\end{align}
The kernel $\Theta(x,y,\underline{\theta})$ is positive-definite with
respect to the chosen training data at $\underline{\theta}$ if,
\footnote{This is stronger than the definition given in \cite{jacotetal18}, which gives
a similar statement for all functions in a given function class $\mathcal{F}$.}
for all $(u_{i})_{i=1}^{N} \in \bigoplus^{N} \mathbb{R}^{Q}$, 
\begin{align}
    \sum_{i=1}^{N} \abs{u_{i}}^{2} >
    0
    \text{ implies } 
    \sum_{i,j=1}^{N} u_{i}^{T} \Theta(x_{i}, x_{j}, \underline{\theta})
    u_{j} > 0 \, .
\end{align}
From \eqref{DDT-ntk} we see that this holds if, and only if, $DD^{T}$ is a
positive-definite matrix.

Thus, Theorem \ref{thm-homotopy} suggests that, by changing the metric in parameter space,
the NTK can be modified into a simpler kernel $K$ which preserves the equilibrium points
of the flow.
When $\Theta$ is positive-definite, $K$ satisfies
\begin{align}
    K(x_{i},x_{j}) = \delta_{ij} \Id_{Q\times Q},
\end{align}
so that it is constant throughout training
at least when evaluated on $\mathcal{X}_{0}$.
This trivializes the training dynamics in output space, and holds for any neural network,
or indeed any family of functions parametrized by $\underline{\theta}$.

\section{Some related work}
\label{sec:relatedwork}

\subsection*{Preconditioners} 
The preconditioner $(D^{T}D)^{+}$ introduced here closely resembles the inverse of the
generalized Gauss--Newton matrix (GN), which approximates the Hessian of the loss with
respect to the parameters:
\begin{align}
    \left(D^{T}D\right)_{ij} &= \,\,\,\, (\partial_{i}\underline{x}^{T}) (\partial_{j}
    \underline{x}), \nonumber \\
    \frac{\partial^{2} \mathcal{C}}{\partial \theta_{i} \partial \theta_{j}} \,\,\, &=
    \underbrace{(\partial_{i}\underline{x}^{T}) H_{\underline{x}} (\partial_{j}
    \underline{x})}_{(GN)_{ij}} +
    \partial_{i} \partial_{j} \underline{x}^{T} \nabla_{\underline{x}}\mathcal{C}.
\end{align}
The distinction lies in omitting $H_{\underline{x}}$, the Hessian with respect to the
outputs, which is the identity for the $L^{2}$ loss. The proof of Theorem
\ref{thm-homotopy} remains
valid for $GN^{-1}$ as long as $H_{\underline{x}}$ is positive-semidefinite in the region
of interest (which holds for the $L^{2}$ and cross-entropy losses), but fails in general. 
Recent work has examined the properties of the GN matrix \cite{zhaosidaklucchi24}, as well
as potential advantages \cite{abreuetal25} of the full Gauss--Newton matrix compared to
Adam \cite{kingmaba17}. For a broader discussion on
curvature-aware optimization, see e.g. \cite{martens20} and references therein.

\subsection*{Linear paths} Some works, starting with 
\cite{goodfellowetal14}, interpolate the initial parameters $\underline{\theta}_{0}$ and
final parameters $\underline{\theta}_{f}$ obtained after training the network with usual
methods such as stochastic gradient descent, 
\begin{align}
    \underline{\theta}_{\alpha} = (1-\alpha )\underline{\theta}_{0} + \alpha
    \underline{\theta}_{f}, \, \alpha\in [0,1],
\end{align}
and study the behavior of the loss function
on these linear slices of parameter space. Recent works, such as \cite{lucasetal21},
provide empirical and theoretical investigations of the monotonic linear interpolation
(MLI) property,
whereby the loss decreases monotonically along the path $\underline{\theta}_{\alpha}$.
This is a different approach from the one taken in this paper, where the linear
interpolation happens in output space.

In \cite{reybelletetal24}, the authors find linear trajectories of probability
distributions when studying generative models and modifying the cost function.

\subsection*{The Jacobian \texorpdfstring{\eqref{jacobian}}{}}
As already discussed in subsection \ref{sec:ntk}, the neural tangent kernel introduced by
\cite{jacotetal18} is closely related to the Jacobian studied in this paper.
It inspired works on linear approximations of gradient flow/descent and the so called
kernel or lazy regime, see e.g. \cite{chizatetal19, woodworthetal20}.

Certain works study the rank of $D$ more explicitly: 
In \cite{grigsbyetal22}, the authors study the functional dimension
of ReLU neural networks, defined in terms of the rank of the Jacobian.
In \cite{karhadkaretal24}, the authors show that for mildly overparametrized ReLU neural
networks with scalar output ($Q=1$) most activation patterns correspond to regions in
parameter space where the Jacobian is full rank.

\vspace{1em}
\noindent
{\bf Acknowledgments:}
T.C. gratefully acknowledges support by the NSF through the grant DMS-2009800, and the RTG
Grant DMS-1840314 - {\em Analysis of PDE}. P.M.E. was supported by NSF grant DMS-2009800
through T.C., and UT Austin's GS Summer fellowship. T.C. and P.M.E. thank the anonymous
referee who suggested we investigate the cross-entropy loss.

\appendix
\section{Pseudoinverse}
\label{appendix}

\renewcommand{\theequation}{\thesection.\arabic{equation}}
\setcounter{equation}{0}

We recall here the definition and properties of the pseudoinverse of a (possibly
rank-deficient) matrix. For proofs and more detail, we refer the reader to
\cite{cm-generalizedinverses}, chapter 1. 
\begin{defi}
    \label{penroseinverse}
    The pseudoinverse of $A \in \mathbb{R}^{m\times n}$ is the unique matrix
    $\pen{A} \in \mathbb{R}^{n \times m}$ such that
    \begin{align}
        A \pen{A} A &= A, \\
        \pen{A} A \pen{A} &= \pen{A}, \\
        (A\pen{A})^{T} &= A\pen{A}, \\
        (\pen{A}A)^{T} &= \pen{A} A.
    \end{align}
\end{defi}
When $A$ is full rank, there are simple expressions for the pseudoinverse: When $\rk(A) =
m = \min\{m,n\}$, $A$ is surjective, so $AA^{T}$ is invertible and we have $\pen{A} =
A^{T}(AA^{T})^{-1}$. On the other hand, when $\rk(A) = n = \min\{m,n\}$, $A$ is injective
and $A^{T}A$ is invertible and we have $\pen{A} = (A^{T}A)^{-1}A^{T}$. In general, there
exist many algorithms to construct the pseudoinverse, for instance using singular value
decomposition; see e.g. \cite{cm-generalizedinverses}, chapter 12.

The matrices $A$ and $\pen{A}$ as above satisfy a few basic properties:
\begin{align}
    \pen{\pen{A}} &= A, \\
    \pen{A^{T}} &= (\pen{A})^{T}, \\
    \pen{A^{T}A} &= \pen{A} \pen{A^{T}}. \label{basicproperty3}
\end{align}

Moreover, the pseudoinverse gives a convenient way to write projectors: Consider a
subspace $V \subset \mathbb{R}^{d}$. We will denote by
$\mathcal{P}_{V}$ the orthogonal projector onto $V$, which satisfies
$\mathcal{P}_{V} = \mathcal{P}_{V}^{2} = \mathcal{P}_{V}^{T}$. Then
\begin{align}
    A\pen{A} &= \mathcal{P}_{\ran(A)}, \label{projector1} \\
    \pen{A} A &= \mathcal{P}_{\ran(A^{T})}.
\end{align}
When considering the subspaces associated to $A$, it is useful to know the relations
\begin{align}
    \ran\pen{A} &= \ran A^{T} , \\
    \ker\pen{A} &= \ker A^{T}. \label{kerproperty}
\end{align}
The following is a useful identity:
\begin{align}
    \label{identity1}
    A = A A^{T} \pen{A^{T}} = \pen{A^{T}} A^{T} A.
\end{align}
Similar identities can be found by substituting $A^{T}$ and $\pen{A}$ in place of $A$.

Finally, note that it is not true in general that $\pen{A}A = A\pen{A}$;
a sufficient condition is $A^{T}=A$. We make use of this fact, along with the following
lemma, in the main text.

\begin{lemma}
    \label{appendix:lemma}
    $\ker(A^{T}A) = \ker(\pen{A}A)$.
\end{lemma}
\begin{proof}
    We expand 
    \begin{align}
        A^{T}A &= A^{T} A \pen{A} A,
    \end{align}
    so
    \begin{align}
        \ker(\pen{A}A) \subset \ker(A^{T}A).
    \end{align}
    Similarly,
    \begin{align}
        \pen{A} A \stackrel{\eqref{identity1}}{=} \pen{A} \pen{A^{T}} A^{T} A
    \end{align}
    and so
    \begin{align}
        \ker(A^{T}A) \subset \ker (\pen{A}A).
    \end{align}
\end{proof}


\begin{thebibliography}{GKRBZ24}

\bibitem[AJVLS17]{ayjost-informationgeometry}
Nihat Ay, J{\"u}rgen Jost, H{\^o}ng V{\^a}n~L{\^e}, and Lorenz Schwachh{\"o}fer.
\newblock {\em Information geometry}, volume~64.
\newblock Springer, 2017.

\bibitem[Ama16]{amari-informationgeometry}
Shun-ichi Amari.
\newblock {\em Information geometry and its applications}, volume 194.
\newblock Springer, 2016.

\bibitem[AVKM25]{abreuetal25}
Natalie Abreu, Nikhil Vyas, Sham Kakade, and Depen Morwani.
\newblock The potential of second-order optimization for llms: A study with full gauss-newton.
\newblock \href{https://arxiv.org/abs/2510.09378}{arXiv:2510.09378}, 2025.

\bibitem[CE25]{chenewald23gd}
Thomas Chen and Patrícia~Muñoz Ewald.
\newblock On non-approximability of zero loss global $\mathcal{L}^2$ minimizers by gradient descent in deep learning.
\newblock {\em Theor. Appl. Mech.}, 52(1):67--73, 2025.
\newblock \href{https://arxiv.org/abs/2311.07065}{arXiv:2311.07065}.

\bibitem[Che23]{chen23}
Thomas Chen.
\newblock Global $\mathcal{L}^2$ minimization at uniform exponential rate via geometrically adapted gradient descent in deep learning, 2023.
\newblock \href{https://arxiv.org/abs/2311.15487}{arXiv:2311.15487}.

\bibitem[CM09]{cm-generalizedinverses}
Stephen~L Campbell and Carl~D Meyer.
\newblock {\em Generalized inverses of linear transformations}.
\newblock SIAM, 2009.

\bibitem[COB19]{chizatetal19}
Lenaic Chizat, Edouard Oyallon, and Francis Bach.
\newblock On lazy training in differentiable programming.
\newblock {\em Advances in Neural Information Processing Systems}, 32, 2019.

\bibitem[Coo21]{cooper21}
Yaim Cooper.
\newblock Global minima of overparameterized neural networks.
\newblock {\em SIAM Journal on Mathematics of Data Science}, 3(2):676--691, 2021.

\bibitem[EW22]{ewojt22}
Weinan E and Stephan Wojtowytsch.
\newblock On the emergence of simplex symmetry in the final and penultimate layers of neural network classifiers.
\newblock In Joan Bruna, Jan Hesthaven, and Lenka Zdeborova, editors, {\em Proceedings of the 2nd Mathematical and Scientific Machine Learning Conference}, volume 145 of {\em Proceedings of Machine Learning Research}, pages 270--290. PMLR, 16--19 Aug 2022.

\bibitem[GKRBZ24]{reybelletetal24}
Hyemin Gu, Markos~A. Katsoulakis, Luc Rey-Bellet, and Benjamin~J. Zhang.
\newblock Combining wasserstein-1 and wasserstein-2 proximals: robust manifold learning via well-posed generative flows, 2024.
\newblock \href{https://arxiv.org/abs/2407.11901}{arXiv:2407.11901}.

\bibitem[GLMW25]{grigsbyetal22}
J~Elisenda Grigsby, Kathryn Lindsey, Robert Meyerhoff, and Chenxi Wu.
\newblock Functional dimension of feedforward relu neural networks.
\newblock {\em Advances in Mathematics}, 482:110636, 2025.

\bibitem[GVS15]{goodfellowetal14}
Ian~J. Goodfellow, Oriol Vinyals, and Andrew~M. Saxe.
\newblock Qualitatively characterizing neural network optimization problems, 2015.
\newblock \href{https://arxiv.org/abs/1412.6544}{arXiv:1412.6544}.

\bibitem[HPD22]{hanpapyandonoho22}
X.Y. Han, Vardan Papyan, and David~L. Donoho.
\newblock Neural collapse under {MSE} loss: Proximity to and dynamics on the central path.
\newblock In {\em International Conference on Learning Representations}, 2022.

\bibitem[JGH18]{jacotetal18}
Arthur Jacot, Franck Gabriel, and Cl{\'e}ment Hongler.
\newblock Neural tangent kernel: Convergence and generalization in neural networks.
\newblock {\em Advances in Neural Information Processing Systems}, 31, 2018.

\bibitem[KB17]{kingmaba17}
Diederik~P. Kingma and Jimmy Ba.
\newblock Adam: A method for stochastic optimization.
\newblock \href{https://arxiv.org/abs/1412.6980}{arXiv:1412.6980}, 2017.

\bibitem[KMTM24]{karhadkaretal24}
Kedar Karhadkar, Michael Murray, Hanna Tseran, and Guido Montúfar.
\newblock Mildly overparameterized relu networks have a favorable loss landscape.
\newblock {\em Transactions on Machine Learning Research}, 2024.
\newblock \href{https://arxiv.org/abs/2305.19510}{arXiv:2305.19510}.

\bibitem[LBZ{\etalchar{+}}21]{lucasetal21}
James~R Lucas, Juhan Bae, Michael~R Zhang, Stanislav Fort, Richard Zemel, and Roger~B Grosse.
\newblock On monotonic linear interpolation of neural network parameters.
\newblock In Marina Meila and Tong Zhang, editors, {\em Proceedings of the 38th International Conference on Machine Learning}, volume 139 of {\em Proceedings of Machine Learning Research}, pages 7168--7179. PMLR, 18--24 Jul 2021.

\bibitem[LM18]{limontufar18}
Wuchen Li and Guido Mont{\'u}far.
\newblock Natural gradient via optimal transport.
\newblock {\em Information Geometry}, 1:181--214, 2018.

\bibitem[Mar20]{martens20}
James Martens.
\newblock New insights and perspectives on the natural gradient method.
\newblock {\em Journal of Machine Learning Research}, 21(146):1--76, 2020.

\bibitem[PHD20]{papyanhandonoho20}
Vardan Papyan, XY~Han, and David~L Donoho.
\newblock Prevalence of neural collapse during the terminal phase of deep learning training.
\newblock {\em Proceedings of the National Academy of Sciences}, 117(40):24652--24663, 2020.

\bibitem[Tes12]{teschlODE}
Gerald Teschl.
\newblock {\em Ordinary differential equations and dynamical systems}, volume 140.
\newblock American Mathematical Soc., 2012.

\bibitem[TKB19]{tragerkohnbruna19}
Matthew Trager, Kathl{\'e}n Kohn, and Joan Bruna.
\newblock Pure and spurious critical points: a geometric study of linear networks.
\newblock {\em arXiv preprint arXiv:1910.01671}, 2019.

\bibitem[WGL{\etalchar{+}}20]{woodworthetal20}
Blake Woodworth, Suriya Gunasekar, Jason~D. Lee, Edward Moroshko, Pedro Savarese, Itay Golan, Daniel Soudry, and Nathan Srebro.
\newblock Kernel and rich regimes in overparametrized models.
\newblock In Jacob Abernethy and Shivani Agarwal, editors, {\em Proceedings of Thirty Third Conference on Learning Theory}, volume 125 of {\em Proceedings of Machine Learning Research}, pages 3635--3673. PMLR, 09--12 Jul 2020.

\bibitem[ZSL24]{zhaosidaklucchi24}
Jim Zhao, Sidak~Pal Singh, and Aurelien Lucchi.
\newblock Theoretical characterisation of the gauss newton conditioning in neural networks.
\newblock {\em Advances in Neural Information Processing Systems}, 37:114965--115000, 2024.

\end{thebibliography}

\newcommand{\etalchar}[1]{$^{#1}$}

\end{document}